\documentclass[11pt,final]{article}

\usepackage{coling10}

\usepackage{times}
\usepackage{latexsym}
\usepackage[fleqn]{amsmath}
\usepackage{amssymb}
\usepackage[utf8x]{inputenc}
\usepackage{t1enc}
\usepackage{graphicx}
\usepackage[T1]{url} 
\usepackage{booktabs}
\usepackage{covington}
\usepackage{sdrt}

\catcode`\=13
\def{$\bowtie$}

\newif\ifpdf
\ifx\pdfoutput\undefined \else
  \ifx\pdfoutput\relax
  \else
    \ifcase\pdfoutput
    \else
      \pdftrue
    \fi
  \fi
\fi

\providecommand{\ie}{\textit{i.e.} }%
\providecommand{\eg}{\textit{e.g.} }%

\providecommand{\citeN}[1]{\newcite{#1}}%

\title{Testing SDRT's Right Frontier}
\author{Stergos D. Afantenos \and Nicholas Asher \\
Institut de recherche en informatique de Toulouse
(IRIT),\\ CNRS, Université Paul Sabatier\\
{\tt \{stergos.afantenos, nicholas.asher\}@irit.fr}}

\begin{document}

\maketitle

\begin{abstract}
The Right Frontier Constraint (\textsc{rfc}), as a constraint on the attachment of new constituents to an existing discourse structure, has important implications for the interpretation of anaphoric elements in discourse and for Machine Learning (\textsc{ml}) approaches to learning discourse structures.  In this paper we provide strong empirical support for \textsc{sdrt}'s version of \textsc{rfc}. The analysis of about 100 doubly annotated documents by five different naive annotators shows that \textsc{sdrt}'s \textsc{rfc}  is respected about 95\% of the time. The qualitative analysis of presumed violations that we have performed shows that they are either click-errors or structural misconceptions.
\end{abstract}

\section{Introduction}
A cognitively plausible way to view the construction of a discourse structure for a text is an incremental one.  Interpreters integrate discourse constituent $n$ into the antecedently constructed discourse structure $D$ for constituents $1$ to $n-1$ by linking $n$ to some constituent in $D$ with a discourse relation.  \textsc{sdrt}'s Right Frontier Constraint (\textsc{rfc}) \cite{Asher.93,Asher:Lascarides.03} says that a new constituent $n$ cannot attach to an arbitrary node in $D$.    Instead it must attach to either the last node entered into the graph or one of the nodes that dominate this last node.  Assuming that the last node is usually found on the right of the structure, this means that the nodes available for attachment occur on the \emph{right frontier} (\textsc{rf}) of the discourse \emph{graph} or \textsc{sdrs}.

Researchers working in different theoretical paradigms have adopted some form of this constraint.  Polanyi \shortcite{Polanyi.85,Polanyi.88} originally proposed the \textsc{rfc} as a constraint on antecedents to anaphoric pronouns.  \textsc{sdrt} generalizes this to a condition on all anaphoric elements.  As the attachment of new information to a contextually given discourse graph in \textsc{sdrt} involves the resolution of an anaphoric dependency, \textsc{rfc} furnishes a constraint on the attachment problem. (Webber, 1988; Mann and Thompson, 1987; 1988)  have also adopted versions of this constraint.  But there are important differences.  While \textsc{sdrt} and \textsc{rst} both take \textsc{rfc} as a constraint on all discourse attachments (in \textsc{dltag}, in contrast, anaphoric discourse particles are not limited to finding an antecedent on the \textsc{rf}), \textsc{sdrt}'s notion of \textsc{rf} is substantially different from that of \textsc{rst}'s or Polanyi's, because \textsc{sdrt}'s notion of a \textsc{rf} depends on a 2-dimensional discourse graph built from {\em coordinating} and {\em subordinating} discourse relations.  Defining {\sc rfc} with respect to SDRT's 2-dimensional graphs allows the \textsc{rf} to contain discourse constituents that do not include the last constituent entered into the graph (in contrast to \textsc{rst}).  \textsc{sdrt} also allows for multiple attachments of a constituent to the \textsc{rfc}.

\textsc{sdrt}'s \textsc{rfc}  has important implications for the interpretation of various types of anaphoric elements:  tense \cite{Lascarides:Asher.93}, ellipsis \cite{Hardt:Asher:Busquets.01,Hardt:Romero.04,Asher.07}, as well as pronouns referring to individuals and abstract entities \cite{Asher.93,Asher:Lascarides.03}. The \textsc{rfc}, we believe, will also benefit \textsc{ml} approaches to learning discourse structures,  as a constraint limiting the search space for possible discourse attachments.  Despite its importance, \textsc{sdrt}'s \textsc{rfc} has never been empirically validated, however. We present evidence in this paper providing strong empirical support for \textsc{sdrt}'s version of the constraint.  We have chosen to study \textsc{sdrt}'s notion of a \textsc{rf}, because of \textsc{sdrt}'s greater expressive power over \textsc{rst} \cite{Danlos.08}, the greater generality of \textsc{sdrt}'s definition of \textsc{rfc}, and because of \textsc{sdrt}'s  greater theoretical reliance on the constraint for making semantic predictions.  \textsc{sdrt} also makes theoretically clear why the \textsc{rfc} should apply to discourse relation attachment, since it treats discourse structure construction as a dynamic process in which all discourse relations are essentially anaphors.  The analysis of about 100 doubly annotated documents by five different naive annotators shows that this constraint, as defined in \textsc{sdrt}, is respected about 95\% of the time. The qualitative analysis of the presumed violations that we have performed shows that they are either click-errors or structural misconceptions by the annotators.

Below, we give a formal definition of \textsc{sdrt}'s \textsc{rfc}; section~\ref{sec:data} explains our annotation procedure. Details of the statistical analysis we have performed are given in section~\ref{sec:expes}, and a qualitative analysis is provided in section~\ref{sec:analysis}. Finally, section~\ref{sec:ml} presents the implications of the empirical study for \textsc{ml} techniques for the extraction of discourse structures while sections~\ref{sec:rw} and~\ref{sec:conc} present the related work and conclusions.

\section{The Right Frontier Constraint in \textsc{sdrt}}\label{sec:rfc_sdrt}

In \textsc{sdrt}, a discourse structure or \textsc{sdrs} (Segmented Discourse Representation Structure) is a tuple $<A, \mathcal{F}, \textsc{last}>$, where $A$ is the set of labels representing the discourse constituents of the structure, $\textsc{last} \in A$ the last introduced label and $\mathcal{F}$ a function which assigns each member of $A$ a well-formed formula of the \textsc{sdrs} language (defined \cite[p~138]{Asher:Lascarides.03}).  \textsc{sdrs}s correspond to $\lambda$ expressions with a continuation style semantics.  \textsc{sdrt} distinguishes coordinating and subordinating discourse relations using a variety of linguistic tests \cite{Asher:Vieu.05},\footnote{The subordinating relations of \textsc{sdrt} are currently: Elaboration (a relation defined in terms of the main eventualities of the related constituents), Entity-Elaboration (E-Elab(a,b) iff b says more about an entity mentioned in a that is not the main eventuality of a) Comment, Flashback (the reverse of Narration), Background, Goal (intentional explanation), Explanation, and Attribution.  The coordinating relations are: Narration, Contrast, Result, Parallel, Continuation, Alternation, and Conditional, all defined in \citeN{Asher:Lascarides.03}.} and isolates structural relations (Parallel and Contrast) based on their semantics.

The \textsc{rf} is the set of available attachment points to which a new utterance can be attached.  What this set includes depends on the discourse relation used to make the attachment.   Here is the definition from \cite[p~148]{Asher:Lascarides.03}.
\begin{quote}\footnotesize
Suppose that a constituent $\beta$ is to be attached to a constituent in the  \textsc{sdrs} with a discourse relation other than Parallel or Contrast.  Then the available attachment points  for $\beta$ are:
\begin{enumerate}
  \item The label $\alpha=\textsc{last}$;
  \item Any label $\gamma$ such that:
      \begin{enumerate}
      \item \textit{i-outscopes}$(\gamma, \alpha)$ (\ie $R(\delta, \alpha)$ or $R(\alpha, \delta)$ is a conjunct in $\mathcal{F}(\gamma)$ for some $R$ and some $\delta$); or
      \item$R(\gamma, \alpha)$ is a conjunct in $\mathcal{F}(\lambda)$ for some label $\lambda$, where $R$ is a subordinating discourse relation.
      \end{enumerate}
      We gloss this as $\alpha < \gamma$.
  \item Transitive Closure:\\
      Any label $\gamma$ that dominates $\alpha$ through a sequence of labels $\gamma_1, \gamma_2, \ldots \gamma_n$ such that $\alpha<\gamma_1<\gamma_2<\ldots\gamma_n< \gamma$
\end{enumerate}
\end{quote}
We can represent an \textsc{sdrs} as a graph $\mathcal{G}$, whose nodes are the labels of the {\sc sdrs}s constituents and whose typed arcs represent the relations between them. The nodes available for attachment of a new element $\beta$ in $\mathcal{G}$ are the last introduced node \textsc{last} and any other node dominating \textsc{last}, where the notion of domination should be understood as the transitive closure over the arrows given by \emph{subordinating} relations or those holding between a complex segment and its parts.  Subordinating relations like \textit{Elaboration} extend the vertical dimension of the graph, whereas coordinating relations like \textit{Narration} expand the structure horizontally. The graph of every {\sc sdrs} has a unique top label for the whole structure or formula; however, there may be multiple $<$ paths defined within a given {\sc sdrs}, allowing for multiple parents, in the terminology of \cite{Wolf:Gibson.06}.  Furthermore, \textsc{sdrt} allows for multiple arcs between constituents and attachments to multiple constituents on the {\sc rfc}, making for a very rich structure.

\textsc{sdrt}'s {\sc rfc} is restricted to non-structural relations, because structural relations postulate a partial isomorphism from the discourse structure of the second constituent to the discourse structure of the first, which provides its own attachment possibilities for subconstituents of the two related structures \cite{Asher.93}.  Sometimes such parallelism or contrast, also known as {\em discourse subordination}  \cite{Asher.93}, can be enforced in a long distance way by repeating the same wording in the two constituents.

{\sc rfc} has the name it does because the segments that belong on this set (the $\gamma$s in the above definition) are typically nodes on a discourse graph which are geometrically placed at the \textsc{rf} of the graph.  Consider the following example embellished from \citeN{Asher:Lascarides.03}:
\begin{examples}\footnotesize
\item\label{ex:1}  (\labone) John had a great evening last night. (\labtwo) He first had a great meal at Michel Sarran. (\labthree) He ate profiterolles de foie gras, (\labfour) which is a specialty of the chef. (\labfive) He had the lobster, (\labsix) which he had been dreaming about for weeks.   (\labseven) He then went out to a several swank bars.
\end{examples}
The graph of the \textsc{sdrs} for \ref{ex:1} looks like this:
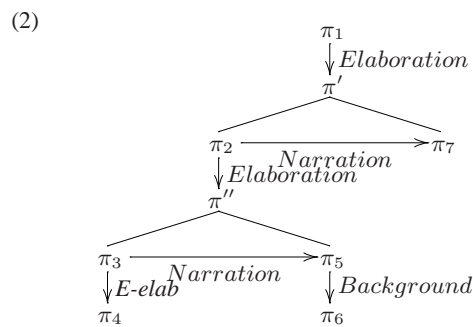
\begin{examples}
\footnotesize
\item \label{tree:1}
\Treek[1]{1}{
  & & \LAB{\labone}\srel{Elaboration} \\
  & & \LAB{\labprime}\consl\consr \\
  & \LAB{\labtwo}\srel{Elaboration}\crel{Narration} & & \LAB{\labseven} \\
  & \LAB{\labsecond}\consl\consr & & \\
  \LAB{\labthree}\srel{\textit{E-elab}}\crel{Narration} & & \LAB{\labfive}\srel{Background}  \\
  \LAB{\labfour} & & \LAB{\labsix}\\
}
\end{examples}
where \labprime\ and \labsecond\ represent complex segments. Given that the last introduced utterance is represented by the node \labseven, the set of nodes that are on the \textsc{rf} are \labseven\ ($\textsc{last}$), \labprime\ (the complex segment that includes \labseven) and \labone\ (connected via a subordinating relation to \labprime). All those nodes are geometrically placed at the \textsc{rf} of the graph.

\textsc{sdrt}'s notion of a \textsc{rf} is more general than \textsc{rst}'s or \textsc{dltag}'s.  First, {\sc sdrs}s can have complex constituents with multiple elements linked by coordinate relations that serve as arguments to other relations, thus permitting instances of {\em shared structure} that are difficult to capture in a pure tree notation \cite{Lee:Prasad:Joshi:Webber.08}.  In addition,  in \textsc{rst} the {\sc rf} picks out the {\em adjacent} constituents, \textsc{last} and complex segments including \textsc{last}.  
Contrary to \textsc{rst}, \textsc{sdrt}, as it uses 2-dimensional graphs, predicts that an available attachment point for \labseven\ is the non local and non adjacent \labtwo, which is distinct from the complex constituent consisting of \labtwo\ to \labsix.\footnote{The 2-dimensionality of {\sc sdrs}s also allows us to represent many examples with Elaboration that involve crossing dependencies in Wolf and Gibson's \shortcite{Wolf:Gibson.06} representation without violation of the {\sc rfc.}}  This difference is crucial to the interpretation of the Narration: Narration claims a sequence of two events; making the complex constituent (essentially a sub-\textsc{sdrs}) an argument of Narration, as \textsc{rst} does, makes it difficult to recover such an interpretation. Danlos's \shortcite{Danlos.08} interpretation of the Nuclearity Principle provides an interpretation of the Narration([2-4],5) that is equivalent to the \textsc{sdrs} graph above.\footnote{\citeN{Baldridge:Asher:Hunter.07}, however, show that the Nuclearity Principle does not always hold.}  But even an optional Nuclearlity Principle interpretation won't help with discourse structures like (\ref{tree:1}) where the backgrounding material in  \labfour\ and the commentary in \labsix\ do \underline{not} and cannot figure as part of the Elaboration for semantic reasons.  In our corpus described below, over 20\% of the attachments were non adjacent; \ie the attachment point for the new material did not include $\textsc{last}$.

A further difference between \textsc{sdrt} and other theories is that, as \textsc{sdrt}'s {\sc rfc}  is applied recursively over complex segments within a given {\sc sdrs}, many more attachment points are available in \textsc{sdrt}.  E.g., consider the {\sc sdrs} for this example, adapted from  \cite{Wolf:Gibson.06}:
\begin{examples}\footnotesize
\item (\labone) Mary wanted garlic and thyme.  (\labtwo) She also needed basil. (\labthree) The recipe called for them.  (\labfour) The basil would be hard to come by this time of year.
\end{examples}
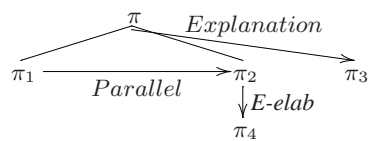
\footnotesize
\Treek[1]{1}{
  & \LAB{$\pi$}\consl\consr\srel[drr]{Explanation} & & \\
\LAB{\labone}\crel{Parallel} & & \LAB{\labtwo}\srel{\textit{E-elab}} & \LAB{\labthree}\\
  & & \LAB{\labfour}& \\
}
\normalsize

Because $\pi$ is the complex segment consisting of $\pi_1$ and $\pi_2$, attachment to $\pi$ with a subordinating discourse relation permits attachment $\pi$'s open constituents as well.\footnote{This part of the {\sc rfc} was not used in \cite{Asher:Lascarides.03}.}

\section{Annotated Corpus}\label{sec:data}
Our corpus comes from the discourse structure annotation project \textsc{annodis}\footnote{\url{http://w3.erss.univ-tlse2.fr/annodis}} which represents an on going effort to build a discourse graph bank for French texts with the two-fold goal of testing various theoretical proposals about discourse structure and providing a seed corpus for learning discourse structures using \textsc{ml} techniques.  \textsc{annodis}'s annotation manual provides detailed instructions about the segmentation of a text into Elementary Discourse Units (\textsc{edu}s).  \textsc{edu}s correspond often to clauses but are also introduced by  frame adverbials,\footnote{Frame adverbials are sentence initial adverbial phrases that can either be temporal, spatial or ``topical" ({\em in Chemistry}).} appositive elements, correlative constructions ({\em [the more you work,] [the more you earn]}), interjections and discourse markers within coordinated VPs {\em [John denied the charges] [\underline{but then} later admitted his guilt]}.  Appositive elements often introduce {\em embedded} \textsc{edu}s; e.g., {\em [Jim Powers, [President of the University of Texas at Austin], resigned today.]}, which makes our segmentation more fine-grained than Wolf and Gibson's \shortcite{Wolf:Gibson.06} or annotation schemes for RST or the PDTB.

The manual also details the meaning of discourse relations but says nothing about the structural postulates of \textsc{sdrt}. For example, there is no mention of the \textsc{rfc} in the manual and very little about hierarchical structure.  Subjects were told to put whatever discourse relations from our list above between constituents they felt were appropriate.  They were also told that they could group constituents together whenever they felt that as a whole they jointly formed the term of a discourse relation. We purposely avoided making the manual too restrictive,  because one of our goals was to examine how well \textsc{sdrt} predicts the discourse structure of subjects who have little knowledge of discourse theories.

In total 5 subjects with little to no knowledge of discourse theories that use \textsc{rfc} participated in the annotation campaign. Three were undergraduate linguistics students and two were graduate linguistics students studying different areas.  The 3 undergraduates benefitted from a completed and revised annotation manual.  The two graduate students did their annotations while the annotation manual was undergoing revisions.  All in all, our annotators doubly annotated about 100 French newspaper texts and {\em Wikipedia} articles. Subjects first segmented each text into \textsc{edu}s, and then they were paired off and compared their segmentations, resolving conflicts on their own or via a supervisor.  The annotation of the discourse relations was performed by each subject working in isolation.  \textsc{annodis} provided a new state of the art tool, \textsc{glozz}, for discourse annotation for the three undergraduates.  With \textsc{glozz} annotators could isolate sections of text corresponding to several \textsc{edu}s, and insert relations between selected constituents using the mouse.   Though it did portray relations selected as lines between parts of the text, \textsc{glozz} did not provide a discourse graph or \textsc{sdrs} as part of its graphical interface.  The representation often yielded a dense number of lines between segments that annotators and evaluators found hard to read.  The inadequate interline spacing in \textsc{glozz} also contributed to certain number of click errors that we detail below in the paper.   The statistics on the number of documents, \textsc{edu}s and relations provided by each annotator are in table~\ref{table:general_stats}.

\begin{table}[htb]
\footnotesize
\centering
  \begin{tabular}{c||c|c|c}
    \toprule
      \textbf{\textit{annotator}} & \textbf{\textit{\# Docs}} & \textbf{\textit{\# \textsc{edu}s}} & \textbf{\textit{\# Relations}}\\
    \midrule
      \textbf{\textit{undergrad 1}} & 27 & 1342 & 1216 \\
      \textbf{\textit{undergrad 2}} & 31 & 1378 & 1302 \\
      \textbf{\textit{undergrad 3}} & 31 & 1376 & 1173  \\
      \textbf{\textit{grad 1}} & 47 & 1387 & 1390 \\
      \textbf{\textit{grad 2}} & 48 & 1314 & 1321   \\
      \bottomrule
 \end{tabular}
 \normalsize
 \caption{Statistics on documents, \textsc{edu}s and Relations.} \label{table:general_stats}
\end{table}

\section{Experiments and Results}\label{sec:expes}
Using \textsc{annodis}'s annotated corpus, we checked  for all \textsc{edu}s $\pi$, whether $\pi$ was attached to a constituent in the \textsc{sdrs} built from the previous \textsc{edu}s in a way that violated the \textsc{rfc}.  Given a discourse as a series of \textsc{edu}s $\pi_1, \pi_2, \ldots, \pi_n$, we constructed for each $\pi_i$ the corresponding sub-graph and calculated the set of nodes on the \textsc{rf} of this sub-graph. We then checked whether the \textsc{edu} $\pi_{i+1}$ was attached to a node that was found in this set.  We also checked whether any newly created complex segment was attached to a node on the \textsc{rf} of this sub-graph.

\subsection{Calculating the Nodes at the \textsc{rf}}
To calculate the nodes on the \textsc{rf}, we slightly extended the annotated graphs, in order to add implied relations left out by the annotators.\footnote{In similar work on TimeML annotations, \citeN{Setzer:Gaizauskas:Hepple.03,Muller:Raymonet.05} add implied relations to annotated, temporal graphs.}

\paragraph{Disconnected Graphs}
While checking the \textsc{rfc} for the attachment of a node $n$, the \textsc{sdrs} graph at this point might consist of 2 or more disjoint subgraphs which get connected together at a later point.
Because we did not want to decide which way these graphs should be connected,  we defined a right frontier for each one using its own $\textsc{last}$.
We then calculated the \textsc{rf} for each one of them and set the set of available nodes to be those in the union of the  \textsc{rf}s of the disjoint subgraphs.  If the subgraphs were not connected at the end of the incremental process in a way that conformed to \textsc{rfc}, we counted this as a violation.   Annotators did not always provide us with a connected graph.  

\paragraph{Postponed Decisions}
\textsc{sdrt} allows for the attachment not only of \textsc{edu}s but also of subgraphs to an available node in the contextually given \textsc{sdrs}. For instance, in the following example, the intended meaning is given by the graph in which the Contrast is between the first label and the complex constituent composed of the disjunction of \labtwo\ and \labthree.
\begin{quote}\footnotesize
  (\labone) Bill doesn't like sports.
  (\labtwo) But Sam does.
  (\labthree) Or John does.
\end{quote}\normalsize

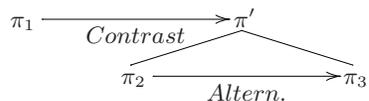
\footnotesize
\Treek[1]{1}{
   \LAB{\labone}\crel{Contrast} & & \LAB{\labprime}\consl\consr  \\
 &   \LAB{\labtwo}\crel{Altern.}&  & \LAB{\labthree} \\
}
\normalsize

Naive annotators attached subgraphs instead of \textsc{edu}s to the \textsc{rf} with some regularity (around 2\%).  This means that an \textsc{edu} $\pi_{i+1}$ could be attached to a node that was not present in the subgraph produced by $\pi_1, \ldots, \pi_i$. There were two main reasons for this:  (1) $\pi_{i+1}$ came from a syntactically fronted clause, a parenthetical or apposition in a sentence whose main clause produced $\pi_{i+2}$ and $\pi_{i+1}$ was attached to $\pi_{i+2}$;  (2) $\pi_{i+1}$ was attached to a complex segment $[\ldots, \pi_{i+1}, \ldots, \pi_{i+k}, \ldots]$ which was not yet introduced in the subgraph.

Since the nodes to which $\pi_{i+1}$ is attached in such cases are not present in the graph, \emph{by definition} they are not in the \textsc{rf} and they could be counted as violations. Nonetheless, if the nodes which connect nodes like $\pi_{i+1}$ eventually link up to the incrementally built \textsc{sdrs} in the right way, $\pi_{i+1}$ might eventually end up linked to something on the \textsc{rf}. For this reason, we postponed the decision on nodes like $\pi_{i+1}$ until the nodes to which they are attached were explicitly introduced in the {\sc sdrs}.

\paragraph{The Coherence of Complex Segments}
In an \textsc{sdrs}, several \textsc{edu}s may combine to form a complex segment $\alpha$ that serves as a term for a discourse relation $R$.  The interpretation of the {\sc sdrs} implies that all of $\alpha$'s constituents contribute to the rhetorical function specified by $R$.  This implies that the coordinating relation \textit{Continuation} holds between the \textsc{edu}s inside $\alpha$, unless there is some other relation between them that is incompatible with Continuation (like a subordinating relation).  Continuations are often used in \textsc{sdrt} \cite{Asher.93,Asher:Lascarides.03}.
%
%
%
During the annotation procedure, our subjects did not always explicitly link the \textsc{edu}s within a complex segment.  In order to enforce the coherence of those complex segments we added \textit{Continuation} relations between the constituents of a complex segment \emph{unless} there was already another path between those constituents.

\paragraph{Expanding Continuations}
Consider the following discourse:
\begin{examples}\footnotesize
\item \label{3}
$[$John, $[$who owns a chain of restau\-rants$]_{\pi_2}$, $[$and is a director of a local charity organization,$]_{\pi_3}$ wanted to sell his yacht.$]_{\pi_1}$ $[$He couldn't afford it anymore.$]_{\pi_4}$
\end{examples}
Annotators sometimes produced the following \textsc{sdrt} graph for the first three \textsc{edu}s of this discourse:
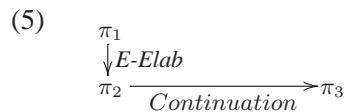
\begin{examples}
\item \label{tree:3}
\footnotesize
\Treek[1]{1}{
  \LAB{\labone}\srel{\textit{E-Elab}} & &   \\
  \LAB{\labtwo}\crel{Continuation} & & \LAB{\labthree} \\
}
\end{examples}
\normalsize

\noindent
In this case the only open node is \labthree\ due to the coordinating relation \textit{Continuation}. Nonetheless, \labfour\ should be attached to \labone, without violating the \textsc{rfc}. Indeed, \textsc{sdrt}'s definition of the Continuation relation enforces that if we have $R(\pi_1,\pi_2)$ and Continuation$(\pi_2,\pi_3)$ then we actually have the complex segment $[\pi_2, \pi_3]$ with $R(\pi_1, [\pi_2, \pi_3])$.  So there is in fact a missing  complex segment in (\ref{tree:3}).  The proper {\sc sdrs} graph of (\ref{3}) is:

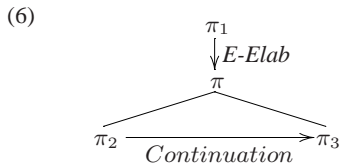
\begin{examples}
\footnotesize
\item \label{tree:3a}
\Treek[1]{1}{
  & \LAB{\labone}\srel{\textit{E-Elab}} &   \\
  & \LAB{$\pi$}\consl\consr & \\
  \LAB{\labtwo}\crel{Continuation} & & \LAB{\labthree} \\
}
\end{examples}
\normalsize

\noindent
which makes \labone\ an available attachment site for \labfour. Such implied constituents have been added to the {\sc sdrs} graphs.

\paragraph{Factoring}
Related to the operation of Expansion, \textsc{sdrt}'s definition of Continuation and various subordinating relations also requires that if we have
$R(a,[\pi_1, \pi_2, \ldots, \pi_n])$ where $[\pi_1, \pi_2, \ldots, \pi_n]$ is a complex segment with $\pi_1, \ldots \pi_n$ linked by Continuation and $R$ is Elaboration, Entity-Elaboration, Frame, Attribution, or Commentary, then we also have $R(a,\pi_i)$ for each $i$.  We added these relations when they were missing.

\subsection{Results}
With the operations just described, we added several inferred relations to the graph.  We then calculated statistics concerning the percentage of attachments for which the \textsc{rfc} is respected using the following formula:
\[RFC_{\mbox{\sc edu}}=\frac{\texttt{\small \# \textsc{edu}s attached to the RF}}{\texttt{\small\# \textsc{edu}s in total}}\]
As we explained, an \textsc{edu} can be attached to an \textsc{sdrt} graph directly by itself or indirectly as part of a bigger complex segment. In order to calculate the nominator we determine first whether an \textsc{edu} directly attaches to the graph's \textsc{rf}, and if that fails we determine whether it is part of a larger complex segment which is attached to the graph's \textsc{rf}. The results obtained are shown in the first two columns of table~\ref{table:rf_stats}. The \textsc{rfc} is respected by at least some attachment decision 95\% of the time--- i.e., 95\% of the \textsc{edu}s get attached to another node that is found on the \textsc{rf}.  The breakdown across our annotators is given in table \ref{table:rf_stats}.

\textsc{sdrt} allows for multiple attachments of an \textsc{edu} to various nodes in an \textsc{sdrs}; \eg while an \textsc{edu} may be attached via one relation to a node on the \textsc{rf}, it may be attached to another node off the \textsc{rf}. To take account of all the attachments for a given \textsc{edu}, we need another way of measuring the percentage of attachments that respects the \textsc{rfc}.  So we counted the ways each \textsc{edu} is related to a node in the \textsc{sdrs} for the previous text and then divided the number of attachment decisions that respect the \textsc{rfc} by the total number of attachment decisions---\ie:\[RFC_r=\frac{\texttt{\footnotesize \# RF attachment decisions}}{\texttt{\footnotesize\# Total attachment decisions}}\].
\begin{table}[htb]
\centering\footnotesize
  \begin{tabular}{c|c|c}
    \toprule
      \textbf{\textit{annotator}} & $RFC_{\mbox{\sc edu}}$ & $RFC_{r}$ \\
      \midrule
      \textbf{\textit{undergrad 1}} & 98.57\% & 91.28\% \\ 
      \textbf{\textit{undergrad 2}} & 98.12\% & 94.39\% \\ 
      \textbf{\textit{undergrad 3}} & 91.93\% & 89.17\% \\ 
      \textbf{\textit{grad 1}} & 94.38\%      & 86.54\% \\ 
      \textbf{\textit{grad 2}} & 92.68\%      & 83.57\% \\ 
      \midrule
      \textit{\textbf{Mean for all annotators}} & 95.24\%& 88.91\% \\
      \textit{\textbf{Mean for 3 undergrad}} & 96.17\% & 91.71\% \\
    \bottomrule
 \end{tabular}
 \normalsize
 \caption{The \% with which each annotator has respected \textsc{sdrt}'s \textsc{rfc} using the \textsc{edu} and attachment decision measures.} \label{table:rf_stats}
\end{table}

The third column of table \ref{table:rf_stats} shows that having a stable annotation manual and \textsc{glozz}  improved the results across our two annotator populations, even though the annotation manual did not say anything about \textsc{rfc} or about the structure of the discourse graphs. Moreover, the distribution of violations of the \textsc{rfc} follows a power law and only 4.56\% of the documents contained more than 5 violations.  This is strong evidence that there is little propagation of violations.

\section{Analysis of Presumed Violations}\label{sec:analysis}
Although 95\% of \textsc{edu}s attach to nodes on the \textsc{rf} of an \textsc{sdrt} graph, 5\% of \textsc{edu}s don't.  \textsc{sdrt} experts performed a qualitative analysis of some of these presumed violations. In many cases, the experts judged that the presumed violations were due to click-errors: sometimes the annotators simply clicked on something that did not translate into a segment.  Sometimes, the experts judged that the annotators picked the wrong segment to attach a new segment or the wrong type of relation during the construction of the \textsc{sdrt} graph.  For example, in the graph that follows the relation between segments 74 and 75 is not a \textit{Comment} but an \textit{Entity-Elaboration}.

As expected, there were also \emph{``structural''} errors, arising from a lack or a misuse of complex segments.  Here is a typical example (translated from the original French):
\begin{quote}\footnotesize
[Around her,]\_74 [we should mention Joseph Racaille]\_75 [responsible for the magnificent arrangements,]\_76 [Christophe Dupouy]\_77 [regular associate of Jean-Louis Murat responsible for mixing,]\_78 [without forgetting her two guardian angels:]\_79 [her agent Olivier Gluzman]\_80 [who signed after a love at first sight,]\_81 [and her husband Mokhtar]\_82 [who has taken care of the family]\_83
\end{quote}

\medskip\noindent\normalsize
Here is the annotated structure up to \textsc{edu} 78:

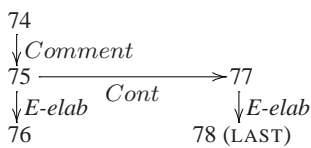
\footnotesize
\Treek[1]{1}{
& \LAB{74}\srel{Comment} & & & \\
& \LAB{75}\srel{\textit{E-elab}}\crel{Cont} & & \LAB{77}\srel{\textit{E-elab}} & & \\
& \LAB{76} & & \LAB{78 (\textsc{last})} & \\
}

%

\normalsize\noindent Note that the attachment of 77 to 75 is non-local and non-adjacent. The annotator then attaches \textsc{edu} 79 to 75 which is blocked from the \textsc{rf} due to the $Continuation$ coordinating relation. By not having created a complex segment due the enumeration that includes \textsc{edu}s 75 to 78, the annotator had no option but to violate the \textsc{rf}. Here is the proper \textsc{sdrt} graph for segments 74 to 79 (where the attachment of 79 to 74 is also both non-local and non-adjacent):

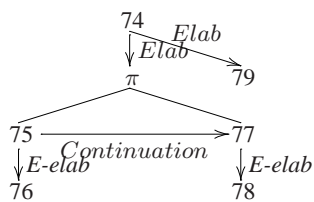
\footnotesize
\Treek[1]{1}{
   &  \LAB{74}\srel{Elab}\srelr{Elab} &     \\
   & \LAB{$\pi$}\consl\consr &      \LAB{79}&  \\
\LAB{75}\srel{\textit{E-elab}}\crel{Continuation} &     & \LAB{77}\srel{\textit{E-elab}} & \\
\LAB{76} &     & \LAB{78} & \\
}

\normalsize\noindent In this case, before the introduction of \textsc{edu} 79, \textsc{edu} 78 is $\textsc{last}$ and by consequence $77, \pi$ and 74 are on the \textsc{rf}. Attaching 79 to 74 is thus legitimate.

We also found more interesting examples of right frontier violations.  One annotator produced a graph for a story which is about the attacks of 9/11/2001 and is too long to quote here. 
A simplified graph of the first part of the story is shown below. \textsc{edu} 4 elaborates on the main event of the story but it is not on the \textsc{rf} for 19.  However,  19 is the first recurrence of the complex definite description {\em le 11 septembre 2001} since the title and the term's definition in \textsc{edu} 4. 

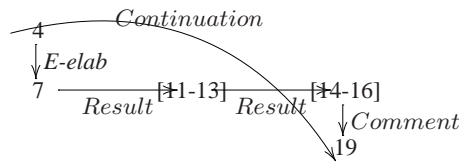
\footnotesize
\Treek[1]{1}{
\LAB{4}\srel{\textit{E-elab}}\Linkk[->]{2}{ddrr} &{Continuation}
&                &                & \\
\LAB{7}\crel[r]{Result} & \LAB{[11-13]}\crel[r]{Result} & \LAB{[14-16]}\srel{Comment} & \\
       &                & \LAB{19} & \\
}\normalsize

\noindent This reuse of the full definite description could be considered a case of \textsc{sdrt}'s discourse subordination.

%
%

\section{\protect\textsc{rfc} and distances of attachment}\label{sec:ml}
Our empirical study vindicates \textsc{sdrt}'s \textsc{rfc}, but it also has computational implications.  Using the \textsc{rfc} dramatically diminishes the number of  attachment possibilities and thus greatly reduces the search space for any incremental discourse parsing algorithm.\footnote{An analogous approach for search space reduction is followed by \citeN{duVerle:Prendinger.09} who use the ``Principle of Sequentiality'' \cite{Marcu2000}, though they do not say how much the search space is reduced.} The mean of nodes that are open on the \textsc{RF} at any given moment on our \textsc{annodis} data is 16.43\% of all the nodes in the graph.%

\begin{figure}[htb]
\setlength{\unitlength}{0.240900pt}
\ifx\plotpoint\undefined\newsavebox{\plotpoint}\fi
\sbox{\plotpoint}{\rule[-0.200pt]{0.400pt}{0.400pt}}%
\begin{picture}(900,540)(0,0)
\sbox{\plotpoint}{\rule[-0.200pt]{0.400pt}{0.400pt}}%
\put(191.0,131.0){\rule[-0.200pt]{4.818pt}{0.400pt}}
\put(171,131){\makebox(0,0)[r]{ 0}}
\put(830.0,131.0){\rule[-0.200pt]{4.818pt}{0.400pt}}
\put(191.0,188.0){\rule[-0.200pt]{4.818pt}{0.400pt}}
\put(171,188){\makebox(0,0)[r]{ 10}}
\put(830.0,188.0){\rule[-0.200pt]{4.818pt}{0.400pt}}
\put(191.0,245.0){\rule[-0.200pt]{4.818pt}{0.400pt}}
\put(171,245){\makebox(0,0)[r]{ 20}}
\put(830.0,245.0){\rule[-0.200pt]{4.818pt}{0.400pt}}
\put(191.0,301.0){\rule[-0.200pt]{4.818pt}{0.400pt}}
\put(171,301){\makebox(0,0)[r]{ 30}}
\put(830.0,301.0){\rule[-0.200pt]{4.818pt}{0.400pt}}
\put(191.0,358.0){\rule[-0.200pt]{4.818pt}{0.400pt}}
\put(171,358){\makebox(0,0)[r]{ 40}}
\put(830.0,358.0){\rule[-0.200pt]{4.818pt}{0.400pt}}
\put(191.0,415.0){\rule[-0.200pt]{4.818pt}{0.400pt}}
\put(171,415){\makebox(0,0)[r]{ 50}}
\put(830.0,415.0){\rule[-0.200pt]{4.818pt}{0.400pt}}
\put(191.0,472.0){\rule[-0.200pt]{4.818pt}{0.400pt}}
\put(171,472){\makebox(0,0)[r]{ 60}}
\put(830.0,472.0){\rule[-0.200pt]{4.818pt}{0.400pt}}
\put(191.0,131.0){\rule[-0.200pt]{0.400pt}{4.818pt}}
\put(191,90){\makebox(0,0){ 0}}
\put(191.0,480.0){\rule[-0.200pt]{0.400pt}{4.818pt}}
\put(257.0,131.0){\rule[-0.200pt]{0.400pt}{4.818pt}}
\put(257,90){\makebox(0,0){ 2}}
\put(257.0,480.0){\rule[-0.200pt]{0.400pt}{4.818pt}}
\put(323.0,131.0){\rule[-0.200pt]{0.400pt}{4.818pt}}
\put(323,90){\makebox(0,0){ 4}}
\put(323.0,480.0){\rule[-0.200pt]{0.400pt}{4.818pt}}
\put(389.0,131.0){\rule[-0.200pt]{0.400pt}{4.818pt}}
\put(389,90){\makebox(0,0){ 6}}
\put(389.0,480.0){\rule[-0.200pt]{0.400pt}{4.818pt}}
\put(455.0,131.0){\rule[-0.200pt]{0.400pt}{4.818pt}}
\put(455,90){\makebox(0,0){ 8}}
\put(455.0,480.0){\rule[-0.200pt]{0.400pt}{4.818pt}}
\put(521.0,131.0){\rule[-0.200pt]{0.400pt}{4.818pt}}
\put(521,90){\makebox(0,0){ 10}}
\put(521.0,480.0){\rule[-0.200pt]{0.400pt}{4.818pt}}
\put(586.0,131.0){\rule[-0.200pt]{0.400pt}{4.818pt}}
\put(586,90){\makebox(0,0){ 12}}
\put(586.0,480.0){\rule[-0.200pt]{0.400pt}{4.818pt}}
\put(652.0,131.0){\rule[-0.200pt]{0.400pt}{4.818pt}}
\put(652,90){\makebox(0,0){ 14}}
\put(652.0,480.0){\rule[-0.200pt]{0.400pt}{4.818pt}}
\put(718.0,131.0){\rule[-0.200pt]{0.400pt}{4.818pt}}
\put(718,90){\makebox(0,0){ 16}}
\put(718.0,480.0){\rule[-0.200pt]{0.400pt}{4.818pt}}
\put(784.0,131.0){\rule[-0.200pt]{0.400pt}{4.818pt}}
\put(784,90){\makebox(0,0){ 18}}
\put(784.0,480.0){\rule[-0.200pt]{0.400pt}{4.818pt}}
\put(850.0,131.0){\rule[-0.200pt]{0.400pt}{4.818pt}}
\put(850,90){\makebox(0,0){ 20}}
\put(850.0,480.0){\rule[-0.200pt]{0.400pt}{4.818pt}}
\put(191.0,131.0){\rule[-0.200pt]{0.400pt}{88.892pt}}
\put(191.0,131.0){\rule[-0.200pt]{158.753pt}{0.400pt}}
\put(850.0,131.0){\rule[-0.200pt]{0.400pt}{88.892pt}}
\put(191.0,500.0){\rule[-0.200pt]{158.753pt}{0.400pt}}
\put(70,430){\makebox(0,0){P}}
\put(70,400){\makebox(0,0){e}}
\put(70,370){\makebox(0,0){r}}
\put(70,340){\makebox(0,0){c}}
\put(70,310){\makebox(0,0){e}}
\put(70,280){\makebox(0,0){n}}
\put(70,250){\makebox(0,0){t}}
\put(70,220){\makebox(0,0){a}}
\put(70,190){\makebox(0,0){g}}
\put(70,150){\makebox(0,0){e}}
\put(520,29){\makebox(0,0){Attachment distance}}
\put(224,473){\usebox{\plotpoint}}
\multiput(224.58,458.04)(0.497,-4.420){63}{\rule{0.120pt}{3.603pt}}
\multiput(223.17,465.52)(33.000,-281.522){2}{\rule{0.400pt}{1.802pt}}
\multiput(257.00,182.92)(1.290,-0.493){23}{\rule{1.115pt}{0.119pt}}
\multiput(257.00,183.17)(30.685,-13.000){2}{\rule{0.558pt}{0.400pt}}
\multiput(290.00,169.92)(0.923,-0.495){33}{\rule{0.833pt}{0.119pt}}
\multiput(290.00,170.17)(31.270,-18.000){2}{\rule{0.417pt}{0.400pt}}
\multiput(323.00,151.93)(2.145,-0.488){13}{\rule{1.750pt}{0.117pt}}
\multiput(323.00,152.17)(29.368,-8.000){2}{\rule{0.875pt}{0.400pt}}
\multiput(356.00,143.94)(4.722,-0.468){5}{\rule{3.400pt}{0.113pt}}
\multiput(356.00,144.17)(25.943,-4.000){2}{\rule{1.700pt}{0.400pt}}
\multiput(389.00,139.95)(7.160,-0.447){3}{\rule{4.500pt}{0.108pt}}
\multiput(389.00,140.17)(23.660,-3.000){2}{\rule{2.250pt}{0.400pt}}
\put(422,136.67){\rule{7.950pt}{0.400pt}}
\multiput(422.00,137.17)(16.500,-1.000){2}{\rule{3.975pt}{0.400pt}}
\put(455,135.67){\rule{7.950pt}{0.400pt}}
\multiput(455.00,136.17)(16.500,-1.000){2}{\rule{3.975pt}{0.400pt}}
\put(553,134.67){\rule{7.950pt}{0.400pt}}
\multiput(553.00,135.17)(16.500,-1.000){2}{\rule{3.975pt}{0.400pt}}
\put(586,133.67){\rule{7.950pt}{0.400pt}}
\multiput(586.00,134.17)(16.500,-1.000){2}{\rule{3.975pt}{0.400pt}}
\put(488.0,136.0){\rule[-0.200pt]{15.658pt}{0.400pt}}
\put(718,132.67){\rule{7.950pt}{0.400pt}}
\multiput(718.00,133.17)(16.500,-1.000){2}{\rule{3.975pt}{0.400pt}}
\put(619.0,134.0){\rule[-0.200pt]{23.849pt}{0.400pt}}
\put(224,473){\raisebox{-.8pt}{\makebox(0,0){$\Diamond$}}}
\put(257,184){\raisebox{-.8pt}{\makebox(0,0){$\Diamond$}}}
\put(290,171){\raisebox{-.8pt}{\makebox(0,0){$\Diamond$}}}
\put(323,153){\raisebox{-.8pt}{\makebox(0,0){$\Diamond$}}}
\put(356,145){\raisebox{-.8pt}{\makebox(0,0){$\Diamond$}}}
\put(389,141){\raisebox{-.8pt}{\makebox(0,0){$\Diamond$}}}
\put(422,138){\raisebox{-.8pt}{\makebox(0,0){$\Diamond$}}}
\put(455,137){\raisebox{-.8pt}{\makebox(0,0){$\Diamond$}}}
\put(488,136){\raisebox{-.8pt}{\makebox(0,0){$\Diamond$}}}
\put(521,136){\raisebox{-.8pt}{\makebox(0,0){$\Diamond$}}}
\put(553,136){\raisebox{-.8pt}{\makebox(0,0){$\Diamond$}}}
\put(586,135){\raisebox{-.8pt}{\makebox(0,0){$\Diamond$}}}
\put(619,134){\raisebox{-.8pt}{\makebox(0,0){$\Diamond$}}}
\put(652,134){\raisebox{-.8pt}{\makebox(0,0){$\Diamond$}}}
\put(685,134){\raisebox{-.8pt}{\makebox(0,0){$\Diamond$}}}
\put(718,134){\raisebox{-.8pt}{\makebox(0,0){$\Diamond$}}}
\put(751,133){\raisebox{-.8pt}{\makebox(0,0){$\Diamond$}}}
\put(784,133){\raisebox{-.8pt}{\makebox(0,0){$\Diamond$}}}
\put(817,133){\raisebox{-.8pt}{\makebox(0,0){$\Diamond$}}}
\put(850,133){\raisebox{-.8pt}{\makebox(0,0){$\Diamond$}}}
\put(751.0,133.0){\rule[-0.200pt]{23.849pt}{0.400pt}}
\put(191.0,131.0){\rule[-0.200pt]{0.400pt}{88.892pt}}
\put(191.0,131.0){\rule[-0.200pt]{158.753pt}{0.400pt}}
\put(850.0,131.0){\rule[-0.200pt]{0.400pt}{88.892pt}}
\put(191.0,500.0){\rule[-0.200pt]{158.753pt}{0.400pt}}
\end{picture}
  \caption{Distribution of attachment distance}\label{fig:distribution}
\end{figure}
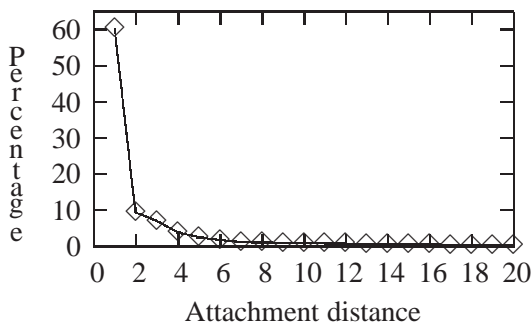

Our data also allowed us to calculate the distance of attachment sites from \textsc{last}, which could be an important constraint on machine learning algorithms for constructing discourse structures.  
Given a pair of constituents $(\pi_i, \pi_j)$ distance is calculated either \emph{textually} (the number of intervening \textsc{edu}s between $\pi_i$ and $\pi_j$) or \emph{topologically} (the length  the shortest path between $\pi_i$ and $\pi_j$).  Topological distance, however, does not take into account the fact that a textually further segment is cognitively less salient. Moreover, this measure can give the same distance to nodes that are textually far away between them due to long distance pop-ups \cite{Asher:Lascarides.03}. A purely textual distance, on the other hand, gives the same distance to an \textsc{edu} $\pi_i$ and a complex segment $[\pi_1, \ldots, \pi_i]$ even if $\pi_1$ and $\pi_i$ are textually distant (since both have the same span end).  We used a measure combining both. The distance scheme that we used assigns to each \textsc{edu} its textual distance from $\textsc{last}$ in the graph under consideration, while a complex segment of rank 1 gets a distance which is computed from the highest distance of their constituent \textsc{edu}s plus 1. For a constituent $\sigma$ of rank $n$ we have:
\[Dist = Max\{\mbox{dist}(x)\colon x \mbox{ in } \sigma \} + n\]
The distribution of attachment follows a power law with 40\% of attachments performed non-locally, that is on segments of distance 2 or more (figure~\ref{fig:distribution}). This implies that the distance between candidate attachment sites that are on the \textsc{rf} is an important feature for an \textsc{ml} algorithm. It is important to note at this point that following the baseline approach of always attaching on the $\textsc{last}$ misses 40\% of attachments.  We also have 20.38\% of the non-local, non-adjacent attachments in our annotations.  So an \textsc{rst} parser using Marcu's \shortcite{Marcu2000} adjacency constraint as do \citeN{duVerle:Prendinger.09} would miss these.

\section{Related Work}\label{sec:rw}
Several studies have shown that the \textsc{rfc} may be violated as an anaphoric constraint when there are other clues, content or linguistic features, that determine the antecedent.  \cite{Poesio:diEugenio.01,Holler:Irmen.07,Asher.08,Prvot:Vieu.08}, for example, show that anaphors such as definite descriptions and complex demonstratives, which often provide enough content on their own to isolate their antecedents, or pronouns in languages like German which must obey gender agreement, might remain felicitous although the discourse relations between them and their antecedents might violate the \textsc{rfc}. 
Usually there are few linguistic clues that help find the appropriate antecedent to a discourse relation, in contrast to the anaphoric expressions mentioned above.  Exceptions involve stylistic devices like direct quotation that license discourse subordination.  Thus, \textsc{sdrt} predicts that \textsc{rfc} violations for discourse attachments should be much more rare than those for the resolution of anaphors that provide linguistic clues about their antecedents.

As regards other empirical validation of various versions of the \textsc{rfc}  for the attachment of discourse constituents, \citeN{Wolf:Gibson.06} show an \textsc{rst}-like \textsc{rfc} is not supported in their corpus GraphBank.  Our study concurs in that some 20\% of the attachments in our corpus cannot be formulated in \textsc{rst}.\footnote{One other study we are aware of is \citeN{Sassen:Khnlein.05}, who show that in chat conversations, the \textsc{rfc} does not always hold unconditionally. Since this genre of discourse is not always coherent, it is expected that the \textsc{rfc} will not always hold here.}  On the other hand, we note that because of the 2 dimensional nature of \textsc{sdrt} graphs and because of the caveats introduced by structural relations and discourse subordination, the counterexamples from GraphBank against, say, \textsc{rst} representations do not carry over straightforwardly to \textsc{sdrs}s.  In fact, once these factors are taken into account, the {\sc rfc} violations in our corpus and in GraphBank are roughly about the same.

\section{Conclusions}\label{sec:conc}
We have shown that \textsc{sdrt}'s \textsc{rfc} has strong empirical support: the attachments of our 3 completely naive annotators fully comply with \textsc{rfc} 91.7\% of the time and partially comply with it  96\% of the time.   As a constraint on discourse parsing \textsc{sdrt}'s {\sc rfc}, we have argued, is both empirically and computationally motivated.   We have also shown that non-local attachments occur about 40\% of the time, which implies that attaching directly on the $\textsc{last}$ will not yield good results.  Further, many of the non local attachments do not respect \textsc{rst}'s adjacency constraint.  We need \textsc{sdrt}'s \textsc{rfc} to get the right attachment points for our corpus.   We believe that empirical studies of the kind we have given here are essential to finding robust and useful features that will vastly improve discourse parsers.


\nocite{Webber.88,Mann&Thompson87,Mann&Thompson88}

\clearpage
\bibliography{coling10_rfc}
\bibliographystyle{coling}
\end{document}